\documentclass[10pt,twoside]{article}

\usepackage{times}
\usepackage[utf8]{inputenc}
\usepackage[T1]{fontenc}
\usepackage{graphicx}
\usepackage{hyperref}
\usepackage[font=small]{caption}

\usepackage{multirow}

\usepackage{LIFT19}
\usepackage[frenchb,english]{babel}

\title{How Does Language Influence Documentation Workflow? Unsupervised Word Discovery Using Translations in Multiple Languages}

\author{Marcely Zanon Boito\up{1}\quad Aline Villavicencio\up{2,3} \quad Laurent Besacier\up{1}\\
  {\small
    (1) Laboratoire d'Informatique de Grenoble (LIG), UGA, G-INP, CNRS, INRIA, France\\ 
    (2) Department of Computer Science, University of Sheffield, England \\ 
    (3) Institute of Informatics (INF), UFRGS, Brazil \\ 
    \texttt{
     \textbf{contact:} marcely.zanon-boito@univ-grenoble-alpes.fr \\ 
}}}

\begin{document}
\maketitle

\resume{Comment la langue influence le processus de documentation~? Découverte non supervisée de mots basée sur des traductions en langues multiples}{
Pour la documentation des langues, la transcription est un processus très coûteux : une minute d'enregistrement nécessiterait environ une heure et demie de travail pour un linguiste \cite{austin2013endangered}. Récemment, la collecte de traductions (dans des langues bien documentées) alignées aux enregistrements est devenue une solution populaire pour garantir l'interprétabilité des enregistrements \cite{adda2016breaking} et aider à leur traitement automatique. Dans cet article, nous étudions l'impact de la langue de traduction sur les approches automatiques en documentation des langues. Nous traduisons un corpus parallèle bilingue Mboshi-Français \cite{godard2017very} dans quatre autres langues, et évaluons l'impact de la langue de traduction sur une tâche de segmentation en mots non supervisée. Nos résultats suggèrent que la langue de traduction peut influencer légèrement la qualité de segmentation. Cependant, combiner l'information apprise par différents modèles bilingues nous permet d'améliorer ces résultats de manière marginale.
}

\abstract{HMMMM}{
For language documentation initiatives, transcription is an expensive resource: one minute of audio is estimated to take one hour and a half on average of a linguist's work~\cite{austin2013endangered}. Recently, collecting aligned translations in well-resourced languages became a popular solution for ensuring posterior interpretability of the recordings~\cite{adda2016breaking}.
In this paper we investigate language-related impact in automatic approaches for computational language documentation. We translate the bilingual Mboshi-French parallel corpus~\cite{godard2017very} into four other languages, and we perform bilingual-rooted unsupervised word discovery. Our results hint towards an impact of the well-resourced language in the quality of the output. However, by combining the information learned by different bilingual models, we are only able to marginally increase the quality of the segmentation.}

\motsClefs
  {découverte non supervisée 
  du lexique, documentation des langues, approches multilingues}
  {unsupervised word discovery, language documentation, multilingual approaches}
 
\section{Introduction}
The \textit{Cambridge Handbook of Endangered Languages}~\cite{austin2011cambridge} estimates that at least half of the 7,000 languages currently spoken worldwide will no longer exist by the end of this century. For these \textit{endangered} languages, data collection campaigns have to accommodate the challenge that many of them are from oral tradition, and producing transcriptions is costly. 
This \textit{transcription bottleneck} problem 
can be handled by translating into a widely spoken language to ensure subsequent interpretability of the collected recordings, and such parallel corpora have been recently created by aligning the collected audio with translations in a well-resourced language~\cite{adda2016breaking, godard2017very, boito2018small}. Moreover, some linguists 
suggested that more than one translation should be collected to capture deeper layers of meaning~\cite{evans2004searching}.


This work is a contribution to the Computational Language Documentation~(CLD) research field, that aims to replace part of the manual steps performed by linguists during language documentation initiatives by automatic approaches. Here we investigate the unsupervised word discovery and segmentation task, using the bilingual-rooted approach from \citet{godard2018unsupervised}. There, words in the well-resourced language are aligned to unsegmented phonemes in the endangered language in order to identify group of phonemes, and to cluster them into word-like units.
We experiment with the Mboshi-French parallel corpus, translating the French text into four other well-resourced languages in order to investigate language impact in this CLD approach. Our results 
hint that this language impact exists, and that models based on different languages will output different word-like units. 

\section{Methodology}
\paragraph{The Multilingual Mboshi Parallel Corpus:}
In this work we extend the bilingual Mboshi-French parallel corpus~\cite{godard2017very}, fruit of the documentation process of Mboshi~(Bantu C25), an endangered language spoken in Congo-Brazzaville. The corpus contains 5,130 utterances, for which it provides audio, transcriptions and translations in French. We translate the French into four other well-resourced languages through the use of the $DeepL$ translator.\footnote{Available at \url{https://www.deepl.com/translator}} The languages added to the dataset are: English, German, Portuguese and Spanish. Table~\ref{tab:stats} shows some statistics for the produced \textit{Multilingual Mboshi} parallel corpus.\footnote{Available at \url{https://github.com/mzboito/mmboshi}}

\paragraph{Bilingual Unsupervised Word Segmentation/Discovery Approach:}
We use the bilingual neural-based Unsupervised Word Segmentation~(UWS) approach from \citet{godard2018unsupervised} to discover words in Mboshi. In this approach, Neural Machine Translation~(NMT) models are trained between language pairs, using as source language the translation (word-level) and as target, the language to document (unsegmented phonemic sequence). Due to the attention mechanism present in these networks~\cite{bahdanau2014neural}, posterior to training, it is possible to retrieve \textit{soft-alignment probability matrices} between source and target sequences. These matrices give us sentence-level source-to-target alignment information, and by using it for clustering neighbor phonemes aligned to the same translation word, we are able to create segmentation in the target side. The product of this approach is a set of (discovered-units, translation words) pairs.

\paragraph{Multilingual Leveraging:}
In this work we apply two simple methods for including multilingual information into the bilingual models from \citet{godard2018unsupervised}.
The first one, \textbf{Multilingual Voting}, consists of 
merging the information learned by models trained with different language pairs by performing a voting over the final discovered boundaries. 
The voting is performed by applying an agreement threshold $T$ over the output boundaries. This threshold balances between accepting all  boundaries from all the bilingual models (zero agreement) and accepting only input boundaries discovered by all these models (total agreement). 
The second method is \textbf{ANE Selection}. For every language pair and aligned sentence in the dataset, a soft-alignment probability matrix is generated. We use \textit{Average Normalized Entropy}~(ANE)~\cite{boito2019empirical} computed over these matrices for selecting \textit{the most confident one} for segmenting each phoneme sequence. This exploits the idea that models trained on different language pairs will have language-related behavior, thus differing on the resulting alignment and segmentation over the same phoneme sequence.

\begin{table}[]
\centering
\includegraphics[scale=0.6]{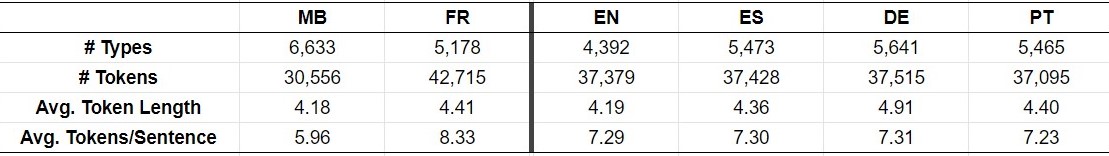}
\caption{Statistics for the Multilingual Mboshi parallel corpus. The French text is used for generating translation in the four other languages present in the right side of the table. 
}
\label{tab:stats}
\end{table}

\section{Experiments}

The experiment settings from this paper and evaluation protocol 
for the Mboshi corpus (Boundary F-scores using the ZRC speech reference) are the same from \citet{boito2019empirical}. Table~\ref{fig:results} presents the results for bilingual UWS and multilingual leveraging. For the former, we reach our best result by using as aligned information the French, the original aligned language for this dataset. Languages closely related to French (Spanish and Portuguese) ranked better, while our worst result used German. English also performs notably well in our experiments. 
We believe this is due to the statistics features of the resulting text. We observe in Table~\ref{tab:stats} that the English portion of the dataset contains the smallest vocabulary among all languages. Since we train our systems in very low-resource settings, vocabulary-related features can impact greatly the system's capacity to language-model, and consequently the final quality of the produced alignments. Even in high-resource settings, it was already attested that some languages are more difficult to model than others~\cite{cotterell2018all}.

\begin{table}
\centering
\includegraphics[scale=0.5]{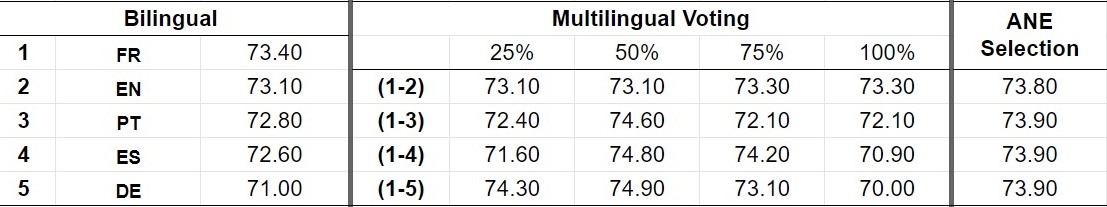}
      \caption{From left to right, results for: 
      bilingual UWS, multilingual leveraging by voting, ANE selection.
      }
      \label{fig:results}
\end{table}

For the multilingual selection experiments, we experimented combining the languages from top to bottom as they appear Table~\ref{fig:results} (ranked by performance; e.g. 1-3 means the combination of FR(1), EN(2) and PT(3)). We observe that the performance improvement is smaller than the one observed in previous work~\cite{boito2019leveraging}, which we attribute to the fact that our dataset was artificially augmented. This could result in the available multilingual form of supervision not being as rich as in a manually generated dataset. Finally, the best boundary segmentation result is obtained by performing multilingual voting with all the languages and an agreement of 50\%, which indicates that the information learned by different languages will provide additional complementary evidence.

Lastly, following the methodology from \citet{boito2019empirical}, we extract the most confident alignments (in terms of ANE) discovered by the bilingual models. Table~\ref{fig:types} presents the top 10 most confident (discovered type, translation) pairs.\footnote{The Mboshi phoneme sequences were replaced by their grapheme equivalents to increase readability, but all results were computed using phonemes.} Looking at the pairs the bilingual models are most confident about, we observe there are some types discovered by all the bilingual models (e.g. Mboshi word \textit{itua}, and the concatenation \textit{oboá+ngá}). However, the models still differ for most of their alignments in the table. This hints that while a portion of the lexicon might be captured independently of the language used, other structures might be more dependent of the chosen language. On this note, \citet{haspelmath2011indeterminacy} suggests the notion of \textit{word} cannot always be meaningfully defined cross-linguistically.

\begin{table}
\centering
\includegraphics[scale=0.6]{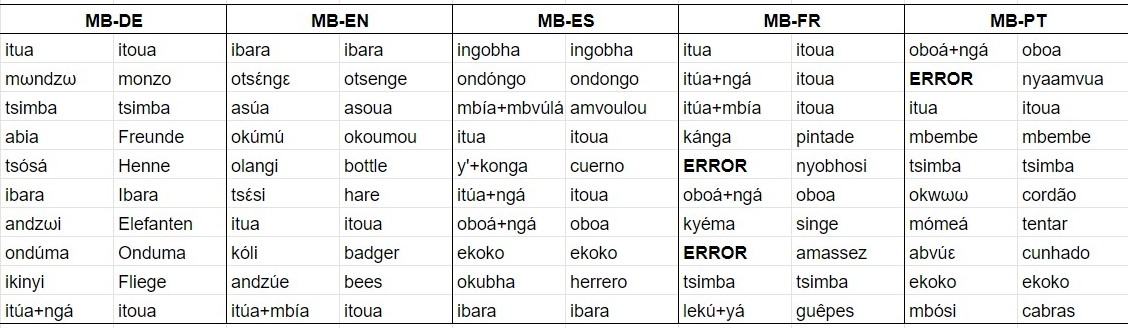}
      \caption{Top 10 confident (discovered type, translation) pairs for the five bilingual models. The ``\textbf{+}'' mark means the discovered type is a concatenation of two existing true types. }
      \label{fig:types}
\end{table}


\section{Conclusion}
In this work we 
train bilingual UWS models using the endangered language Mboshi as target and different well-resourced languages as aligned information. 
Results show that similar languages rank better in terms of segmentation performance, 
and that 
by combining the information learned by different models, segmentation is further improved. 
This might be due to the different language-dependent structures that 
are captured by using more than one language. Lastly, we extend the bilingual Mboshi-French parallel corpus, creating a multilingual corpus for the endangered language Mboshi that we make available to the community.


\bibliographystyle{apalike}
\bibliography{bibliolift19}

\end{document}